%% file: main.tex
\documentclass[10pt,twocolumn,letterpaper]{article}

\usepackage[pagenumbers]{cvpr} 

\input{preamble}

\definecolor{cvprblue}{rgb}{0.21,0.49,0.74}
\usepackage[pagebackref,breaklinks,colorlinks,allcolors=cvprblue]{hyperref}

\usepackage{multirow}
\usepackage{booktabs}
\usepackage[utf8]{inputenc} 
\usepackage[T1]{fontenc}    
\usepackage{hyperref}       
\usepackage{url}            
\usepackage{amsfonts}       
\usepackage{nicefrac}       
\usepackage{microtype}      
\usepackage{xcolor}         
\usepackage{graphicx}
\usepackage{tcolorbox}
\usepackage{xcolor}
\usepackage{adjustbox}
\usepackage[skip=2pt,font=footnotesize]{caption}
\usepackage{array}
\usepackage{widetable}
\usepackage[inline]{enumitem}
\usepackage{amsmath}
\usepackage{pifont}
\usepackage{wrapfig}
\usepackage{xspace}
\usepackage{bbm}
\usepackage{bm}

\usepackage{etoolbox}
\AtBeginDocument{%
  \setlength{\abovedisplayskip}{6pt}%
  \setlength{\belowdisplayskip}{6pt}%
  \setlength{\abovedisplayshortskip}{4pt}%
  \setlength{\belowdisplayshortskip}{4pt}%
}

\usepackage[margin=1in]{geometry}
\usepackage{amsmath, amssymb, amsthm}
\usepackage{mathtools}
\usepackage{bm}
\usepackage{graphicx}
\usepackage{booktabs}
\usepackage{microtype}
\usepackage{enumitem}
\usepackage{xcolor}
\usepackage{xspace}
\usepackage{algpseudocode}
\usepackage{siunitx}

\usepackage[ruled,vlined]{algorithm2e}
\SetAlCapNameFnt{\normalfont}        
\SetAlCapFnt{\normalfont}            

\usepackage{soul}
\newcommand{\hlgreen}[1]{\sethlcolor{green!20}\hl{#1}}
\newcommand{\hlred}[1]{\sethlcolor{red!15}\hl{#1}}

\def\paperID{35070} 
\def\confName{CVPR}
\def\confYear{2026}

\title{Decoupling Vision and Language: Codebook Anchored Visual Adaptation}

\author{
Jason Wu$^{1,2}$\thanks{Work conducted during an internship at AWS AI Lab.}\quad
Tianchen Zhao$^{1}$\quad
Chang Liu$^{1}$\quad
Jiarui Cai$^{1}$\quad
Zheng Zhang$^{1}$\\
Zhuowei Li$^{1}$\quad
Aaditya Singh$^{1}$\quad
Xiang Xu$^{1}$\quad
Mani Srivastava$^{1,2}$\quad
Jonathan Wu$^{1}$\\[0.5em]
$^{1}$AWS\quad
$^{2}$University of California, Los Angeles\quad
}

\begin{document}
\maketitle
\input{sec/0_abstract}    
\input{sec/1_intro}

\input{sec/2_related}

\input{sec/3_method}

\input{sec/4_experiment}

{
    \small
    \bibliographystyle{ieeenat_fullname}
    \bibliography{main}
}

\input{sec/X_suppl}

\end{document}

%% file: preamble.tex
\usepackage{etoolbox}
\makeatletter
\apptocmd{\@maketitle}{\vspace{-11pt}}{}{}
\makeatother

%% file: sec/0_abstract.tex
\begin{abstract}
Large Vision–Language Models (LVLMs) use their vision encoders to translate images into representations for downstream reasoning, but the encoders often underperform in domain-specific visual tasks such as medical image diagnosis or fine-grained classification, where representation errors can cascade through the language model, leading to incorrect responses.
Existing adaptation methods modify the continuous feature interface between encoder and language model through projector tuning or other parameter-efficient updates, which still couples the two components and requires re-alignment whenever the encoder changes.
We introduce CRAFT (Codebook RegulAted Fine-Tuning), a lightweight method that fine-tunes the encoder using a discrete codebook that anchors visual representations to a stable token space, achieving domain adaptation without modifying other parts of the model.
This decoupled design allows the adapted encoder to seamlessly boost the performance of LVLMs with different language architectures, as long as they share the same codebook. Empirically, CRAFT achieves an average gain of 13.51\% across 10 domain-specific benchmarks such as VQARAD and PlantVillage, while preserving the LLM’s linguistic capabilities and outperforming peer methods that operate on continuous tokens.
\end{abstract}

%% file: sec/1_intro.tex
\section{Introduction}
\label{sec:intro}

The vision encoders of LVLMs often struggle on tail domains or tasks that were underrepresented during pretraining~\cite{zhang2024whyvlmsbad, yuksekgonul2022bags, liu2025perceptionbottleneck}. They may misinterpret important visual cues or overlook domain-specific details required for the task, and these errors can propagate through the alignment layers into the language model, misleading the entire LVLM towards incorrect answers, \textit{e.g.}, underperforming human accuracy by more than 50\% on curated evaluations that stress vision-encoder weaknesses~\cite{tong2024eyeswideshut}. The common strategy is to perform fine-tuning with domain-specific data. However, updating the encoder alone is rarely sufficient: once its feature distribution shifts, the language model must also be recalibrated to interpret the new visual embeddings. This implies that each time a new domain is introduced or a stronger language backbone is adopted, the alignment process has to be repeated from the beginning.

\begin{figure}[t]
    \centering
    \includegraphics[width=0.75\linewidth]{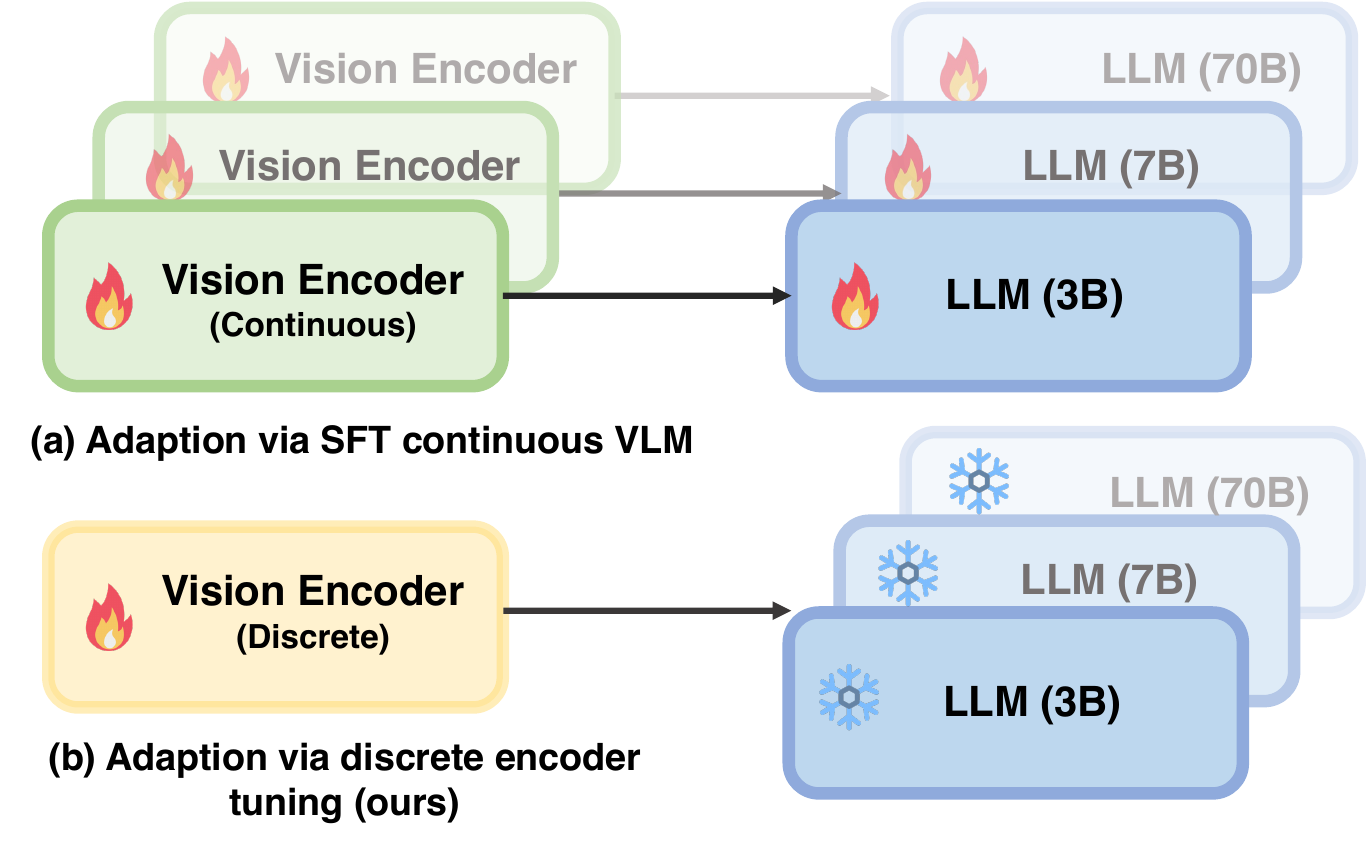}
    \caption{Continuous \textit{vs.} Discrete Adaptation. (a) In conventional continuous-space adaptation, fine-tuning the vision encoder shifts its feature distribution, requiring costly re-alignment with each language model.
    (b) CRAFT introduces a discrete interface that anchors visual features to a shared codebook, allowing a single adapted encoder to work seamlessly across language models of different architectures without additional re-training or alignment.}
    \label{fig:teaser_0}
\end{figure}

The more fundamental problems arise when both the vision encoder and the language model are adapted together.
On the data side, training under narrowly scoped, domain-specific supervision makes the model prone to memorizing superficial text patterns and can thus forget its linguistic capabilities; as noted in DeepSeek-VL~\citep{lu2024deepseek}, this necessitates a carefully curated mixture of multimodal data, which is challenging to obtain in practice. On the compute side, adapting the full multimodal stack quickly becomes impractical because even small changes in the vision encoder force the entire language model to be retrained -- a process that can involve billions of parameters and is seldom necessary for the target task. These constraints motivate a fundamental question: \textit{Can we adapt a large vision–language model without ever touching the original LLM?}

\begin{figure*}[t]
    \centering
    \includegraphics[width=\linewidth]{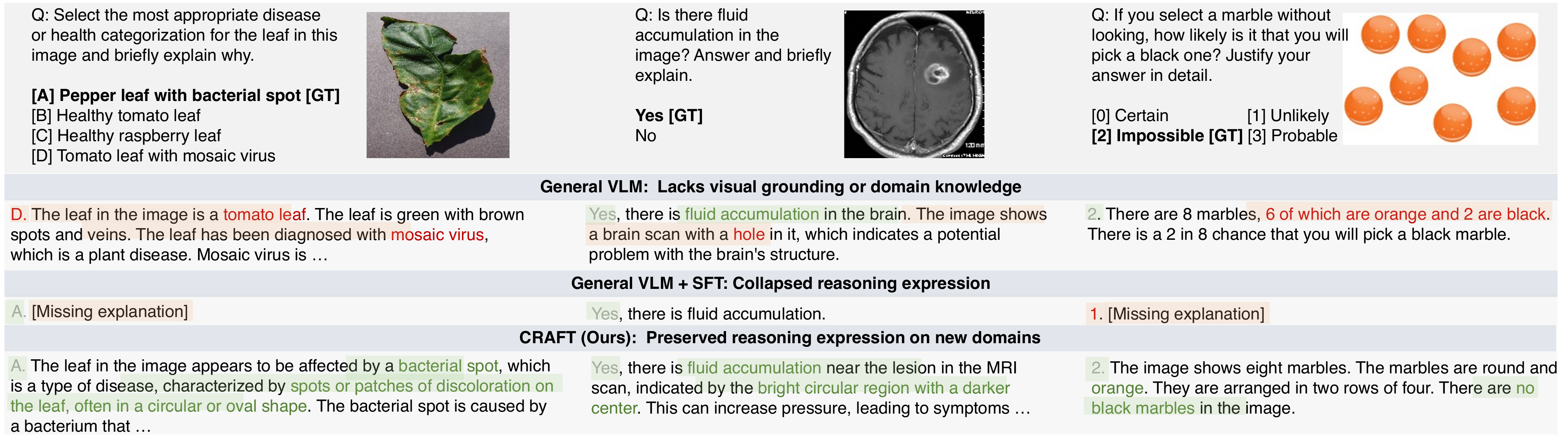}
    \caption{
        Examples from plant pathology~\cite{salathe2016plantvillage}, medical imaging~\cite{lau2018dataset}, and abstract diagram understanding~\cite{lu2021iconqa} are shown using a general continuous LVLM~\cite{lin2023vila}, its PEFT-tuned variant, and our CRAFT model built on the discrete LVLM~\cite{wu2024vila}. General LVLM often lacks visual grounding or domain-specific knowledge in under-represented domains (\textit{e.g.}, misidentifying plant diseases). PEFT improves task accuracy such as question answering, but its language output collapses into rigid responses. In contrast, CRAFT captures domain-specific visual cues (\textit{e.g.}, identifying lesions in medical images) while keeping alignment stable through the shared discrete token interface, allowing the model to produce both accurate decisions and coherent explanations. Correct and incorrect answers or explanations are marked in \hlgreen{green} and \hlred{red}, respectively.
    }
    \label{fig:teaser}
    \vspace{-2mm}
\end{figure*}

Achieving this goal requires bridging the representational gap between a frozen language model and an evolving vision encoder.
The key challenge is maintaining consistent LLM comprehension of the new visual embedding space, while ensuring that the encoder’s updates genuinely enhance multimodal reasoning rather than distort cross-modal alignment.
Inspired by recent advances in discretized LVLMs~\cite{wu2024vila, qu2025tokenflow} which demonstrate on-par or even superior performance against their continuous counterparts, we propose \textbf{CRAFT} (\textbf{C}odebook \textbf{R}egul\textbf{A}ted \textbf{F}ine-\textbf{T}uning), a lightweight framework for vision encoder adaptation in discrete LVLMs that builds upon a shared ``world language'' between visual and textual modalities. 
Specifically, CRAFT discretizes continuous visual embeddings according to a shared, frozen codebook before feeding them into the language model. Conceptually, adapting the discrete vision encoder to a new domain involves learning how to select and arrange existing codebook entries so that they carry the visual evidence the language model needs to solve the target tasks. Once trained, the encoder speaks a stable “visual vocabulary’’ defined by the codebook and can be plugged into any LVLM that uses the same codebook, effectively improving in-domain performance without retraining the language component, as illustrated in Figure~\ref{fig:teaser_0}.

To make these tokens both faithful to the image and concise for reasoning, CRAFT adopts a composite training objective with a test-time regularizer. During training, a surrogate language model scores the joint image-text sequence and backpropagates the loss to the visual encoder, guiding it to select discrete tokens that are genuinely useful for the target domain. At test time, a discrete token pruning scheme removes redundant, background-like tokens and preserves the informative ones, so the LLM receives a compact visual summary instead of a cluttered grid of patches.
This design is lightweight in both computation and data: training cost is mainly determined by the size of the surrogate model, which can be much smaller than the inference backbone, and the method requires only domain-specific supervision without curating extra instruction data to prevent forgetting. Yet, CRAFT outperforms both continuous-feature baselines and their PEFT counterparts, offering sharper domain-specific understanding without sacrificing the model’s ability to follow instructions. As shown in Figure~\ref{fig:teaser}, the vanilla model misinterprets a bright region in a brain scan as a ``hole,’’ while a LoRA fine-tuned model collapses to short answers even when explanations are requested. In contrast, the CRAFT encoder learns to describe the region as a ``bright circular region with a darker center,’’ giving the frozen LLM enough signal to recognize fluid accumulation while preserving full reasoning capability. Our contributions are summarized as follows:

\begin{itemize}
    \item We introduce CRAFT, a lightweight framework for adapting LVLMs that fine-tunes only the discrete vision encoder while keeping the language model frozen, enabling the adapted encoder to transfer across LLM backbones that share the same codebook.

    \item We develop a training and inference scheme that combines surrogate-based supervision and a test-time token pruning strategy, which together improve visual inputs to the LLM with domain-specific priors.

    \item Across ten benchmarks, CRAFT improves domain-specific performance by an average of 13.51\% points while preserving strong instruction-following and explanatory abilities, outperforming continuous-feature and PEFT-based baselines.
\end{itemize}

\begin{figure*}[t!]
\centering
\includegraphics[width=1.0\linewidth]{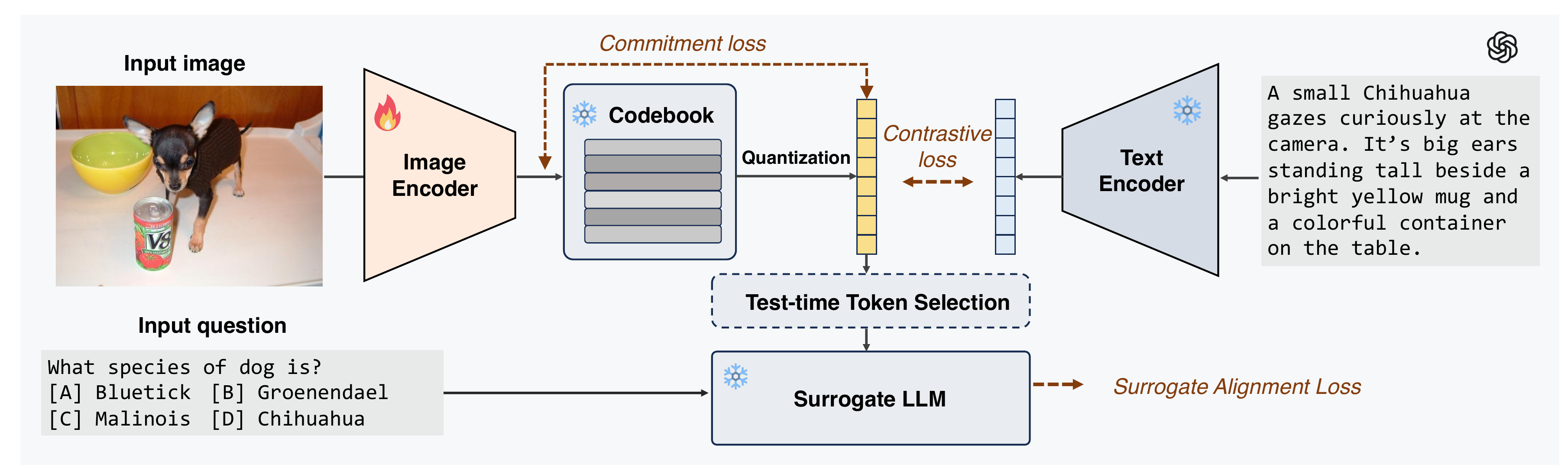}
\caption{Overview of the CRAFT framework. CRAFT adapts to a target domain by fine-tuning only the vision encoder, and its outputs are quantized into a shared discrete codebook. Training is guided by commitment and contrastive losses, with an additional surrogate LLM providing multimodal supervision. At inference, the adapted encoder can be used with any LLM that shares the same codebook.}
\label{fig:main_method}
\vspace{-4mm}
\end{figure*}

%% file: sec/2_related.tex
\section{Related Work}
\label{sec:related}

\noindent\textbf{Domain Adaptation.}
Improving general-purpose LVLMs for specific domains through fine-tuning is an active research area~\citep{zhang2024visually, geigle2024african, he2025analyzing, li2025fakescope, li2025llava, liu2024pefomed, chen2024efficiency}, mostly performing supervised instruction fine-tuning~\citep{chung2024scaling} on domain-specific text–image datasets. Current methods usually update only the projector~\citep{zhang2024visually, geigle2024african}, a small number of LLM parameters~\citep{chen2024efficiency}, or use LoRA-based fine-tuning~\citep{hu2022lora, li2025fakescope, li2025llava, liu2024pefomed}. However, these methods can cause the model to forget its original instruction-following ability~\citep{ren2024analyzing}, even when using parameter-efficient approaches~\citep{lin2024lora}, if the new data are not well balanced. In addition, tuning the LLM or projector cannot fix perceptual errors from the vision encoder. For example, \citet{zhang2024visually} found that errors from the visual backbone can lead to unrecoverable mistakes in the LVLM during fine-grained classification. In contrast, we propose a simple and lightweight method to adapt domains using expert discrete vision encoders, which can be directly plugged into LLMs that share the same visual codebook, improving in-domain performance without inducing catastrophic forgetting.

\noindent\textbf{Adapted Vision Encoders.} 
LVLMs usually rely on CLIP-style vision encoders to produce visual embeddings that align with text to serve as input to the LLM~\citep{bai2025qwen2, li2024llavaonevision}.
Although domain-specific fine-tuning of vision encoders is still underexplored, some studies have used extra vision encoders~\citep{shen2024mome, lin2023sphinx, jiang2023clip} or scaled existing ones~\citep{fan2025scalinglf, chen2023internvl} to boost general performance.
Other works have fine-tuned CLIP encoders outside of LVLMs~\citep{han2024anchor, mukhoti2023fine, ypsilantis2025infusing, sanyal2025upweighting, zhao2025salient}.
For example, \citet{mukhoti2023fine} applied contrastive fine-tuning with an $\ell_2$ regularization term to mitigate forgetting.
These studies show that visual embeddings play a crucial role in LVLM quality, motivating us to focus on fine-tuning the vision encoder for domain-specific improvement.

\noindent\textbf{Discrete LVLMs.}
Discrete LVLMs have gained popularity by unifying text and image generation in a single framework~\citep{jin2023unified, wu2024vila, qu2025tokenflow, lin2025toklip, xie2025muse}. A major challenge is balancing the vision encoder’s ability to both \textit{understand} and \textit{generate} visual content~\citep{wang2024emu3, team2024chameleon, zhou2024transfusion, zhao2025qlip, chen2025semhitok, xie2025show, lin2025toklip, song2025dualtoken}, as these tasks rely on different levels of visual abstraction. In this work, we demonstrate another advantage of discrete vision tokens over continuous ones: they allow seamless interchange between expert vision encoders and LLM backbones that share a common visual codebook.

%% file: sec/3_method.tex
\section{Methodology}

Adapting the discrete vision encoder of an LVLM to a target domain is essentially learning how to select a subset of visual tokens from a shared codebook that best supports the language model in solving domain-specific tasks. The goal is to make the selected tokens both faithful to the image and concise in representation, so that they convey the most relevant visual information for downstream multimodal reasoning. To achieve this, CRAFT introduces a training objective and a test-time regularization scheme. During training, we fine-tune the vision encoder with a composite loss that guides it to produce tokens that are most helpful for the target domain when interpreted by the language model. At test time, we apply discrete vision token pruning to remove redundant tokens, improving the clarity and focus of the visual input. The two stages form a lightweight and modular approach to domain adaptation through a shared discrete interface, transferring domain knowledge to the LVLM without changing the language model component.

\subsection{Preliminaries}
\label{sec:prelim}

We denote the visual encoder as $E_\theta$ and a frozen codebook as $\mathcal{C}=\{c_k\in\mathbb{R}^d\}_{k=1}^{K}$.  
Given an image $x$, the encoder produces a grid of continuous features 
$z = E_\theta(x)\in\mathbb{R}^{N\times d}$, where each vector $z_i$ corresponds to one of the $N$ local patches.  
In a \textit{continuous} latent model, these feature vectors are directly projected into the language model’s embedding space.
In a \textit{discrete} latent model, each $z_i$ is replaced by its nearest entry in the codebook:
\begin{align}
\tilde{z}_i = \text{q}(z_i) = c_{i^*}, \quad 
i^* = \arg\min_{k}\|z_i - c_k\|_2,
\end{align}
where q(·) denotes quantization.
This operation maps the continuous feature grid $z$ to a sequence of discrete visual tokens 
$\tilde{z} = (\tilde{z}_1, \dots, \tilde{z}_N)$, 
which can also be represented by their corresponding codebook indices 
$(i_1, i_2, \dots, i_N)$, where each $i_n$ identifies a visual token $c_{i_n}\in\mathcal{C}$. This idea was first introduced by VQVAE~\citep{van2017neural} and later refined in discrete vision–language models such as VQGAN~\citep{esser2021taming} and VILA-U~\citep{wu2024vila}, with the latter extending the codebook into a hierarchical form that captures both coarse and fine visual semantics. The token embeddings $\tilde{z}$ are then mapped into the LLM’s embedding space by a projector $g_\phi:\mathbb{R}^d\!\to\!\mathbb{R}^{d_{\text{LM}}}$ and concatenated with text tokens for autoregressive processing by the language model.

\subsection{Training Process}
\label{sec:training}

As shown in Figure~\ref{fig:main_method}, CRAFT adapts the vision encoder with three training objectives. 
The \textit{surrogate alignment loss} updates the encoder through next-token prediction using a surrogate language model, providing end-to-end supervision from text. 
The \textit{commitment loss} keeps the encoder’s outputs close to their assigned codebook entries. 
The \textit{contrastive loss} regularizes the encoder to preserve the semantic structure learned during pretraining, maintaining the overall quality of visual representations. 
Each objective is introduced in more detail below.

\noindent\textbf{Surrogate Alignment Loss.}
The goal of adapting a vision encoder to a new domain is not just to make it ``see'' the right pixels, but to generate discrete tokens that capture domain-specific cues interpretable by a language model. Standard visual objectives can refine image features, but they do not teach the encoder which details are actually useful for question answering. As illustrated in Figure~\ref{fig:teaser}, the vanilla model mistakes a bright region in a brain scan for a ``hole,'' an interpretation that might seem plausible in general settings but is clearly invalid in medical imaging. This happens because the vision encoder passes low-level signals to the LLM that do not convey meaningful context. The surrogate alignment loss addresses this issue by letting a surrogate language model evaluate the joint image-text sequence and propagate gradients back through the encoder. It guides the encoder to select and arrange codebook tokens that the language model can naturally reason over for the target task. 
Note that the surrogate language model serves only as a bridge during training and is not used at test time, as our goal is to train the encoder to produce tokens that capture domain-relevant visual cues for reasoning, regardless of which LLM later interprets them.

Specifically, the surrogate alignment loss $\mathcal{L}_{\text{SAL}}$ is defined as an autoregressive loss computed using the surrogate language model $\mathcal{M}$:
\begin{align}
  \mathcal{L}_{\text{SAL}} =
  - \sum_{i=1}^{|y|}
  \log p_{\mathcal{M}}\!\left(y_i \mid y_{<i},\, t,\, \tilde{z} \right),
\end{align}
where $\tilde{z} = \text{q}\!\left(E_\theta(x)\right)$, and $x$, $t$, and $y$ represent the input image, input prompt, and target output, respectively.

\noindent\textbf{Commitment Loss.}
During training, the encoder produces continuous features that are mapped to their nearest codebook entries. Without regularization, these features can drift away from the assigned entries, weakening the connection between the encoder output and the codebook. As a result, the quantized tokens may no longer reflect the visual input faithfully. To address this, we apply a \textit{commitment loss} that keeps the encoder outputs close to the codebook entries:
\begin{equation}
\begin{aligned}
\mathcal{L}_{\text{commit}} = \big\| E_\theta(x) - \operatorname{sg}\!\big[\text{q}\big(E_\theta(x)\big)\big] \big\|_2^2.
\end{aligned}
\end{equation}
Here, $\text{q}(\cdot)$ denotes quantization and $\operatorname{sg}(\cdot)$ is the stop-gradient operator. Note that unlike standard vector quantization methods~\cite{van2017neural}, which also update the codebook during training, our codebook remains fixed, and the encoder is trained with the commitment loss that keeps its features well represented after quantization.

\noindent\textbf{Contrastive Loss.}
Following common practice in the pre-training of discrete LVLMs, we further adopt a contrastive objective that leverages both image captions produced by captioning models and ground-truth text refined by an LLM.
More specifically, for classification tasks, each caption is expanded with: \textit{``An image of a \{domain\}, specifically a \{label\}''}. For VQA tasks, each question--answer pair is rewritten as a declarative statement, \textit{e.g.}, \textit{``Q: What color is the car? A: Red''} becomes \textit{``The car is red''}. 
These augmented captions help connect the image to both its visual content and the intended task.
Each image $x_i$ is paired with a set of text samples: $\mathcal{T}_i = \{ t_i^{\text{GT}} \} \cup \{ t_i^{(1)}, t_i^{(2)}, \ldots, t_i^{(L_i)} \}$, where $t_i^{\text{GT}}$ is the ground-truth text and $t_i^{(l)}$ are generated captions~\citep{li2023blip,zhu2023minigpt}.
At each training step, we sample one sentence $\tilde{t}_i \in \mathcal{T}_i$ and encode it using a frozen text encoder to obtain $\mathbf{t}_i \in \mathbb{R}^{d_{\text{LM}}}$.
The contrastive loss~\citep{zhai2023sigmoid} is formulated as:
\begin{align}
\mathcal{L}_{\text{con}}^{i}
= -\log \sigma(s_{ii})
  - \sum_{j: y_j \neq y_i}
    \log \big(1 - \sigma(s_{ij})\big),
\end{align}
where $\sigma(\cdot)$ is the sigmoid function, $s_{ij} = \tau \cos(\mathbf{v}_i, \mathbf{t}_j)$ is the cosine similarity between the image embedding $\mathbf{v}_i$ and text embedding $\mathbf{t}_j$, $\tau$ is a learnable temperature parameter, and $y_i$ is the class label of image $x_i$.

CRAFT combines three complementary components into a single training objective:
\begin{align}
    \mathcal{L}_{\text{CRAFT}} = 
    \lambda_{\text{con}} \mathcal{L}_{\text{con}} +
    \lambda_{\text{commit}} \mathcal{L}_{\text{commit}} +
    \mathcal{L}_{\text{SAL}} .
\end{align}
During optimization, only the vision encoder $E_\theta$ is updated with this loss; the surrogate language model and the projector remain frozen.
Since the $\arg\min$ in Eq. (1) is non-differentiable, we use the straight-through estimator to propagate gradients through the quantization step during training. For the contrastive loss, we prepend a learnable [CLS] token to input image token sequence and use its pre-quantization embedding from the penultimate encoder layer as the image-level representation $\mathbf{v}_i$.

\subsection{Test-Time Vision Token Pruning}

\begin{figure}[t]
\centering
\includegraphics[width=1.0\linewidth]{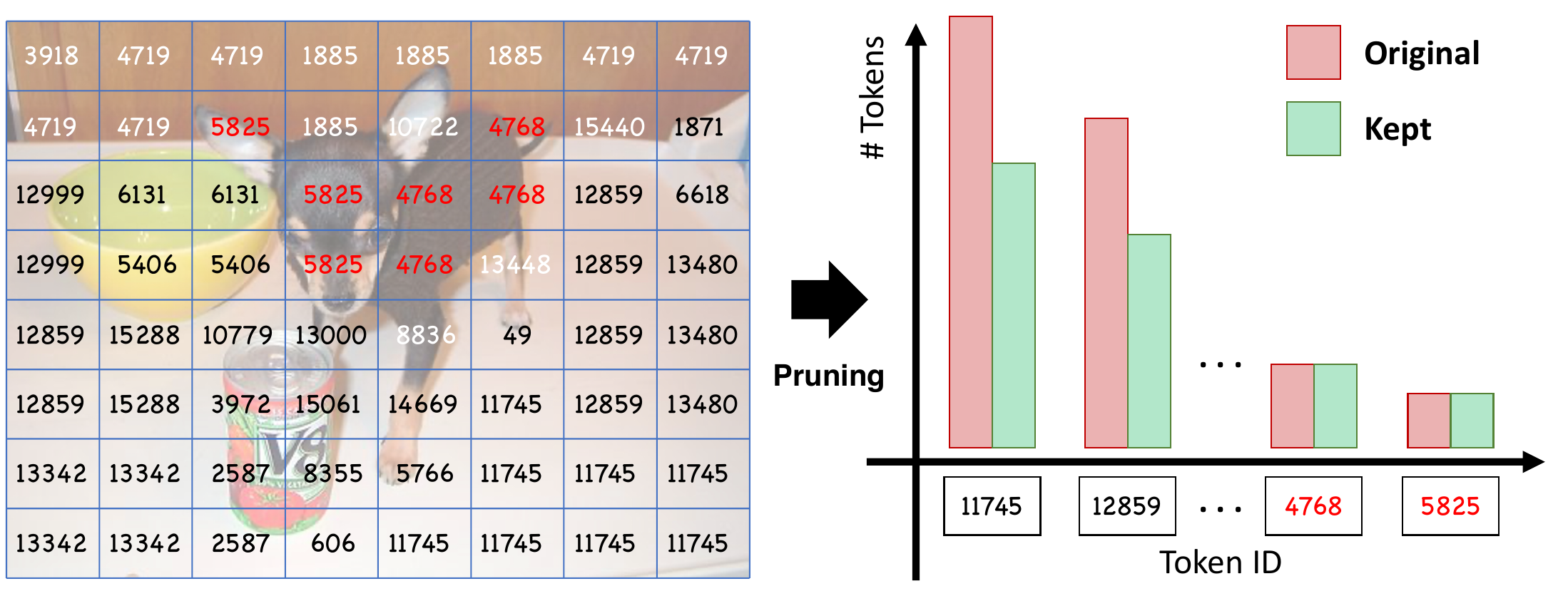}
\caption{Illustration of the Token Pruning Process. Most of the white background patches are mapped to the same codebook entry (ID~11745), whereas semantically meaningful objects such as the Chihuahua are represented by rarer token IDs (ID-5825). Token IDs that appear too often in the training set usually repeat information already captured by others. To reduce this redundancy, we regularize them by pruning a subset at test-time. Tokens that appear frequently during training receive lower rarity weights and are pruned more aggressively than those with higher rarity weights.}
\label{fig:prune}
\end{figure}

As shown in Figure~\ref{fig:prune}, discrete vision encoders transform an image into a grid of codebook token indices, each corresponding to a local patch. 
However, not all tokens are equally informative for downstream reasoning, \textit{e.g.}, frequent background tokens carry redundant information and can be pruned with little semantic loss.
We introduce a simple test-time discrete visual token pruning method that adaptively removes redundant tokens, allowing the model to focus on salient concepts of the image, thereby improving its performance.

\noindent\textbf{Rarity-weighted allocation.}
Let $n_k$ denote the number of tokens in the current image that are assigned to codebook entry $k$, 
with $\sum_k n_k = N$. 
From the training set, we estimate the global frequency $p_{\text{dom}}(k)$ for each entry, 
and define a rarity weight $\rho_k = 1 / p_{\text{dom}}(k)$. We allocate a per-ID keep quota $m_k$ according to
$m_k = \left\lceil n_k \, \rho_k^{\,\gamma} \right\rceil$,
where $\gamma$ controls the overall pruning strength. 
To satisfy a target budget $M < N$, we solve $\gamma$ via a one-dimensional search so that the total number of kept tokens matches the budget:
$
\sum_k \left\lceil n_k \, \rho_k^{\,\gamma} \right\rceil \approx M.
$
Here, setting $\gamma = 0$ keeps all tokens, and increasing $\gamma$ gradually prunes more redundant ones until the target budget is satisfied.

\noindent\textbf{Within-ID selection.}
For each codebook entry $k$, exactly $m_k$ of its $n_k$ tokens are kept based on two cues. 
First, we favor tokens with large quantization residuals 
$e_j = \| z_j - c_k \|_2^2$ because they are harder to quantize and typically correspond to regions that are underrepresented during training~\citep{neloy2024comprehensive}. 
Second, we prefer spatially isolated tokens by computing each token's average distance 
to its nearest neighbors of the same codebook ID and keeping those with the largest values. 
This encourages diverse spatial coverage and prunes densely clustered background patches.

\begin{table*}[t]
\centering
\setlength{\tabcolsep}{4.5pt}
\renewcommand{\arraystretch}{1.15}
\begin{adjustbox}{max width=1.0\linewidth}
\begin{tabular}{l|l|l|l|cccccccccc|c}
\hline
Method & Train Surro. & Inference BB. & Vision Token & IconQA & OCRVQA & ScienceQA & VQARAD & EuroSAT & Flowers & Kvasir & PlantVillage & Cars & Dogs & Avg. \\
\hline
Zero-shot & - & VILA-U-7B & Discrete & 10.17 & 66.40 & 65.59 & 35.67 & 69.15 & 75.80 & 29.20 & 43.83 & 72.50 & 82.40 & 55.07 \\
Zero-shot & - & VILA-7B & Continuous & 31.27 & 59.20 & 68.81 & 41.67 & 79.48 & 76.96 & 39.63 & 47.20 & 76.30 & 78.33 & 59.89 \\
\cline{1-15}
Vision FT & VILA-7B & VILA-7B & Continuous & 39.97 & 62.27 & 67.42 & 43.67 & 67.35 & 79.02 & 37.54 & 62.13 & 86.80 & 71.43 & 61.76 \\
Projector~\citep{zhang2024visually} & VILA-7B & VILA-7B & Continuous & 34.47 & 57.37 & 67.91 & 42.67 & 65.61 & 63.03 & 41.36 & 49.00 & 75.57 & 69.03 & 56.60 \\
LDIFS~\citep{mukhoti2023fine} & VILA-7B & VILA-7B & Continuous & 28.97 & 59.31 & 67.18 & 41.92 & 63.20 & 63.70 & 36.93 & 48.54 & 74.79 & 65.30 & 54.98 \\
\multirow{5}{*}{\textbf{Ours}} & Qwen2-0.5B & VILA-U-7B & Discrete & 30.50 & 66.77 & 68.86 & 39.00 & 78.87 & 72.31 & 32.55 & 70.70 & 78.80 & 74.11 & 61.25 \\
 & Qwen2.5-0.5B & VILA-U-7B & Discrete & 31.90 & 65.90 & 68.81 & 42.33 & 75.76 & 72.18 & 29.65 & 57.53 & 76.7 & 72.01 & 59.28 \\
 & Qwen2-1.5B & VILA-U-7B & Discrete & 33.30 & 66.80 & \textbf{69.15} & 42.33 & \textbf{83.11} & 74.10 & 29.19 & 83.43 & 77.54 & 77.94 & 63.69 \\
 & Qwen2.5-3B & VILA-U-7B & Discrete & 33.73 & 67.97 & 69.06 & \textbf{46.33} & 79.70 & 77.92 & 33.71 & 66.70 & 87.34 & 83.81 & 64.63 \\
 & VILA-U-7B & VILA-U-7B & Discrete & \textbf{48.50} & \textbf{68.13} & 68.71  & 45.67 & 77.80 & \textbf{80.26} & \textbf{41.93} & \textbf{77.27} & \textbf{92.74} & \textbf{84.77} & \textbf{68.58} \\
\hline
\end{tabular}
\end{adjustbox}
\caption{Comparison of zero-shot and fine-tuned models across 10 classification and VQA datasets. We report accuracy (\%) using the exact match criterion. Discrete vision models are trained with different surrogates sharing the same codebook and evaluated on the VILA-U-7B backbone, achieving decent performance gains over the VILA-U-7B baseline. Furthermore, our method outperforms peer approaches that focus on vision encoder fine-tuning. For a fair comparison, we only consider contemporary language model families such as Llama 2, Qwen 2, and Qwen 2.5.}
\label{tab:zero-vs-finetuned-full}
\vspace{-3mm}
\end{table*}

\begin{table}[t]
\centering
\small
\setlength{\tabcolsep}{4pt}
\begin{adjustbox}{max width=1.0\linewidth}
\begin{tabular}{l|cccccc}
\toprule
Model & Corr.\,$\uparrow$ & Pres.\,$\uparrow$ & Rel.\,$\uparrow$ & Faith.\,$\uparrow$ & Verb.\,$\downarrow$ & Overall\,$\uparrow$ \\
\midrule
\multicolumn{7}{c}{\textit{VQA-RAD}} \\[-2pt]
\cmidrule(lr){1-7}
VILA-U-7B                               & 33.34 & 38.31 & 1.44 & 1.08 & 3.03 & 1.01 \\
VILA-7B                                 & 42.99 & \textbf{82.34} & \textbf{3.08} & 1.79 & 3.41 & 3.17 \\
VILA-LDIFS~\citep{mukhoti2023fine}      & 46.31 & 24.35 & 0.69 & 0.40 & 2.66 & -0.24 \\
VILA-Projector FT~\citep{zhang2024visually}  & 44.89 & 4.01  & 0.28 & 0.22 & \textbf{2.21} & -0.61 \\
VILA-LLM-LoRA~\citep{hu2022lora}        & 44.65 & 6.34  & 0.26 & 0.25 & 2.97 & -0.98 \\
\textbf{Ours}                                  & \textbf{47.34} & 75.98 & 2.95 & \textbf{1.99} & 3.47 & \textbf{3.21} \\
\midrule
\multicolumn{7}{c}{\textit{Flowers}} \\[-2pt]
\cmidrule(lr){1-7}
VILA-U-7B                                 & 78.48 & 22.68 & 0.74 & 0.68 & 2.50 & 0.17 \\
VILA-7B                                   & 80.16 & 44.03 & 1.92 & \textbf{1.81} & 2.33 & \textbf{2.57} \\
VILA-LDIFS~\citep{mukhoti2023fine}        & 81.31 & 18.01 & 0.82 & 0.62 & 2.56 & 0.16 \\
VILA-Projector FT~\citep{zhang2024visually} & 69.98 & 20.51 & 0.74 & 0.66 & 2.97 & -0.09 \\
VILA-LLM-LoRA~\citep{hu2022lora}          & \textbf{84.69} & 2.18  & 0.08 & 0.04 & \textbf{1.92} & -0.84 \\
\textbf{Ours}                                     & 81.65 & \textbf{48.65} & \textbf{2.09} & 1.75 & 2.79 & 2.45 \\
\midrule
\multicolumn{7}{c}{\textit{Cars}} \\[-2pt]
\cmidrule(lr){1-7}
VILA-U-7B                                 & 74.85 & 38.34 & 1.71 & 1.54 & 2.60 & 1.95 \\
VILA-7B                                   & 80.15 & \textbf{46.18} & 1.55 & 1.29 & 2.88 & 1.40 \\
VILA-LDIFS~\citep{mukhoti2023fine}        & 86.89 & 10.32 & 0.01 & 0.01 & \textbf{1.85} & -0.91 \\
VILA-Projector FT~\citep{zhang2024visually} & 76.68 & 21.98 & 0.62 & 0.51 & 2.12 & 0.07 \\
VILA-LLM-LoRA~\citep{hu2022lora}          & \textbf{94.18} & 0.01  & 0.02 & 0.02 & \textbf{1.85} & -0.89 \\
\textbf{Ours}                                     & 92.68 & 41.98 & \textbf{1.99} & \textbf{1.60} & 2.81 & \textbf{2.19} \\
\midrule
\multicolumn{7}{c}{\textit{Dogs}} \\[-2pt]
\cmidrule(lr){1-7}
VILA-U-7B                                 & 83.13 & 22.33 & 1.03 & 0.88 & 3.08 & 0.37 \\
VILA-7B                                   & 78.34 & 26.66 & 1.24 & 1.03 & 2.93 & 0.81 \\
VILA-LDIFS~\citep{mukhoti2023fine}        & 73.66 & 4.37  & 0.26 & 0.19 & 2.55 & -0.83 \\
VILA-Projector FT~\citep{zhang2024visually} & 71.48 & 7.02  & 0.64 & 0.35 & 2.09 & -0.05 \\
VILA-LLM-LoRA~\citep{hu2022lora}          & \textbf{88.02} & 2.37  & 0.09 & 0.05 & \textbf{1.77} & -0.75 \\
\textbf{Ours}                                     & 85.13 & \textbf{38.65} & \textbf{1.39} & \textbf{1.13} & 2.66 & \textbf{1.19} \\
\bottomrule
\end{tabular}
\end{adjustbox}
\caption{Comparison of reasoning performance across four datasets. Each row reports correctness (Corr), explanation presence (Pres), relevance (Rel), faithfulness (Faith), verbosity (Verb; lower is better), and the combined Overall score. Zero-shot VILA-U-7B and VILA-7B achieve moderate correctness but provide weaker explanations. Continuous fine-tuning methods, especially LLM-LoRA, often raise correctness but severely degrade instruction-following and explanation quality, leading to low Pres, Rel, Faith, and Overall scores. CRAFT offers a better balance, matching or improving both correctness and explanation presence and quality over both VILA and its continuously fine-tuned variants, achieving an overall better performance.}
\label{tab:explain_summary}
\end{table}

\noindent\textbf{Inference-time pruning.} Collecting all selected tokens gives the final subset $S^{\star}$ with $|S^{\star}| \approx M$. We sort these tokens in their original raster order and feed them into the projector and language model. We define the \textit{keep ratio} as $M/N$, meaning that roughly $N-M$ tokens are pruned. This pruning strategy produces a compact and semantically faithful visual representation that improves efficiency and robustness of LVLM.

%% file: sec/4_experiment.tex
\begin{table*}[t]
\centering
\small
\setlength{\tabcolsep}{4.5pt}
\renewcommand{\arraystretch}{1.15}
\begin{adjustbox}{max width=\linewidth}
\begin{tabular}{l|l|l|cccccccccc|c}
\hline
Method & Train Surrogate & Inference Backbone & IconQA & OCRVQA & ScienceQA & VQARAD & EuroSAT & Flowers & Kvasir & PlantVillage & Cars & Dogs & Avg. \\
\hline
Zero-shot & - & Qwen2-0.5B & 2.40 & 55.43 & 43.43 & 36.67 & 61.70 & 42.84 & 28.74 & 28.23 & 35.47 & 35.17 & 37.01 \\
\textbf{Ours} & Qwen2-0.5B & Qwen2-0.5B & \textbf{30.97} & 54.33 & 47.24 & \textbf{39.00} & \textbf{80.68} & \textbf{52.29} & \textbf{38.11} & \textbf{76.43} & 57.94 & \textbf{67.17} & \textbf{54.42} \\
\textbf{Ours} & VILA-U-7B & Qwen2-0.5B & 17.73 & \textbf{60.27} & \textbf{47.44} & 37.33 & 70.70 & 51.44 & 31.85 & 69.67 & \textbf{57.70} & 54.77 & 49.89 \\
\hline
Zero-shot & - & Qwen2.5-0.5B & 1.57 & 41.93 & 55.63 & 36.67 & 67.46 & 50.89 & 32.68 & 27.63 & 31.37 & 34.60 & 38.04 \\
\textbf{Ours} & Qwen2-0.5B & Qwen2.5-0.5B & \textbf{26.13} & 43.70 & \textbf{56.91} & \textbf{42.67} & \textbf{81.46} & \textbf{61.88} & \textbf{46.68} & \textbf{70.27} & \textbf{50.10} & \textbf{65.54} & \textbf{54.53} \\
\textbf{Ours} & VILA-U-7B & Qwen2.5-0.5B & 19.27 & \textbf{51.73} & \textbf{56.91} & 41.00 & 70.26 & 57.10 & 43.67 & 65.97 & 43.97 & 57.97 & 50.79 \\
\hline
Zero-shot & - & Qwen2-1.5B & 7.07 & 59.33 & 68.27 & 37.00 & 68.20 & 55.15 & 33.26 & 51.17 & 55.27 & 55.90 & 49.06 \\
\textbf{Ours} & Qwen2-0.5B & Qwen2-1.5B & \textbf{22.93} & 59.47 & 74.11 & \textbf{42.67} & \textbf{82.87} & \textbf{69.48} & \textbf{44.37} & \textbf{82.53} & 73.54 & \textbf{80.54} & 63.25 \\
\textbf{Ours} & VILA-U-7B & Qwen2-1.5B & 22.90 & \textbf{68.67} & \textbf{74.41} & 41.33 & 80.68 & 68.44 & 34.29 & 81.37 & \textbf{85.87} & 78.24 & \textbf{63.62} \\
\hline
Zero-shot & - & Qwen2.5-3B & 3.80 & 58.13 & 63.26 & 33.67 & 66.46 & 59.16 & 22.25 & 46.73 & 63.07 & 50.90 & 46.74 \\
\textbf{Ours} & Qwen2-0.5B & Qwen2.5-3B & 19.80 & 59.63 & 63.60 & 36.33 & \textbf{84.11} & 65.76 & \textbf{47.15} & \textbf{81.97} & 70.60 & \textbf{70.84} & 59.98 \\
\textbf{Ours} & VILA-U-7B & Qwen2.5-3B & \textbf{25.03} & \textbf{69.17} & \textbf{63.70} & \textbf{39.67} & 80.50 & \textbf{66.02} & 38.69 & 81.37 & \textbf{77.64} & 67.41 & \textbf{60.92} \\
\hline
Zero-shot & - & VILA-U-7B & 10.17 & 66.40 & 65.59 & 35.67 & 69.15 & 75.80 & 29.20 & 43.83 & 72.50 & 82.40 & 55.07 \\
\textbf{Ours} & Qwen2-0.5B & VILA-U-7B & 30.50 & 66.77 & \textbf{68.86} & 39.00 & \textbf{78.87} & 72.31 & 32.55 & 70.70 & 78.80 & 74.11 & 61.25 \\
\textbf{Ours} & VILA-U-7B & VILA-U-7B & \textbf{48.50} & \textbf{68.13} & 68.71 & \textbf{45.67} & 77.80 & \textbf{80.26} & \textbf{41.93} & \textbf{77.27} & \textbf{92.74} & \textbf{84.77} & \textbf{68.58} \\
\hline
\end{tabular}
\end{adjustbox}
\caption{Cross-LLM transfer with a shared discrete codebook. A vision encoder trained with one surrogate language model during training can be directly paired with different LLM backbones at test time. The adapted encoder consistently improves the performance of five LLM backbones across ten datasets, demonstrating that CRAFT supports portable vision encoder fine-tuning without requiring re-alignment of language models.}
\label{tab:crossllm}
\vspace{-2mm}
\end{table*}

\section{Experiments}
\label{sec:results}

\subsection{Experimental Setup}

\noindent\textbf{Tasks and Metrics.}
We evaluate VQA accuracy and multimodal reasoning using a unified protocol. We consider the following benchmarks: \textit{IconQA}~\citep{lu2021iconqa}, \textit{OCRVQA}~\citep{mishraICDAR19}, \textit{ScienceQA}~\citep{lu2022learn}, \textit{VQARAD}~\citep{lau2018dataset}, \textit{EuroSAT}~\citep{helber2019eurosat}, \textit{Flowers}~\citep{nilsback2006visual}, \textit{Kvasir}~\citep{Pogorelov:2017:KMI:3083187.3083212}, \textit{PlantVillage}~\citep{salathe2016plantvillage}, \textit{Cars}~\citep{krause20133d}, and \textit{Dogs}~\citep{khosla2011novel}. For classification tasks (\textit{e.g.}, \textit{Dogs}, \textit{Cars}, \textit{Flowers}), we cast them to multiple choice question answering format with four options. In addition, we randomly remove 20\% of label categories during training to measure out-of-domain generalization. All datasets use short-answer formats. For accuracy evaluation, we cap decoding at 5 tokens and compute exact-match accuracy after normalizing case and whitespace. For reasoning evaluation, the model may generate up to 1024 tokens. To ensure reproducibility, we fix random seeds across all runs and report the average over 3 seeds. Unless mentioned otherwise, we use a keep ratio of 0.8. See Appendix for more details.

\noindent\textbf{Language Model Backbones.}
We conduct experiments using two LLM families: Llama2-7B (used in VILA-U-7B and VILA-7B) and Qwen2/2.5 (0.5B, 1.5B, 3B), each paired with a SigLIP-Large (ViT-L/16@256) visual encoder. To build surrogates for our experimental setting, we follow the training procedure outlined in the VILA-U paper and train four Qwen-based LLM backbones to be compatible with the VILA-U vision codebook. More details are provided in the Appendix.

\noindent\textbf{Training Details.}
Unless stated otherwise, we train the model with a global batch size of 24 using the AdamW optimizer. We apply a cosine learning rate schedule with a 3\% warmup ratio, no weight decay. During fine-tuning, only the vision tower is updated, the projector and the language model remain frozen. We set $\lambda_{\text{commit}}$=0.1 and $\lambda_{\text{con}}$ to be 0.1 for VQA datasets and 1.0 for visual grounding datasets like \textit{Cats} and \textit{Dogs}, \textit{etc.}

\subsection{Performance Comparison}
We compare CRAFT with the following vision encoder tuning methods. In the \textit{Zero-shot} setting, the model is tested without any fine-tuning. In the \textit{Vision FT} setting, we update only the continuous image encoder and keep both the projector and the language model frozen, using an autoregressive loss. In the \textit{Projector FT} setting, following Zhang et al.~\citep{zhang2024visually}, we adjust only the projector weights, a method mainly studied for classification tasks. We also consider LDIFS~\cite{mukhoti2023fine}, which reduces CLIP feature drift by adding an $\ell_2$ regularization term. In later experiments, we compare against LLM LoRA~\cite{hu2022lora}, where the language model is fine-tuned using LoRA adapters.

The results are shown in Table \ref{tab:zero-vs-finetuned-full}. We first compare CRAFT with the zero-shot baselines with VILA and VILA-U, and find that CRAFT delivers large improvements across five surrogate models and ten datasets. For example, fine-tuning the vision encoder with only a 0.5B surrogate LLM boosts \textit{PlantVillage} and \textit{IconQA} accuracy by 26.87\% and 20.33\%, respectively. In the strongest setting, where both the surrogate and the inference backbone are 7B models, CRAFT surpasses existing continuous-feature fine-tuning approaches by 6.82\% on average, even though it is built on discrete models, which are often regarded as less competitive on standard benchmarks.

However, the gains are not consistent across datasets. One possible reason is that some domains are limited by the reasoning ability of the language model, such as \textit{ScienceQA}, which reduces the benefit of having stronger visual features. Another factor is the capacity gap between the surrogate model and the inference backbone. On fine-grained datasets like \textit{Flowers} and \textit{Dogs}, using a small surrogate such as Qwen2-0.5B lowers accuracy to 72.31\% and 74.11\%, compared with the zero-shot accuracy of 75.80\% and 82.40\%. As shown in Table~\ref{tab:crossllm}, the surrogate itself performs poorly (42.84\% and 35.17\%), suggesting that a weak surrogate may not provide the encoder with useful signals to learn features beyond what a strong backbone already supports. In contrast, using a stronger surrogate consistently improves performance, giving gains of 4.46\% on \textit{Flowers} and 2.37\% on \textit{Dogs}.

\subsection{Evaluation of Instruction-Following}
\label{sec:exp:explainability}

CRAFT does not need to re-align the LLM to accommodate the distribution shift of vision features. This not only reduces training cost but also improves data efficiency. For example, fine-tuning an LLM on datasets with short answers can cause it to forget how to follow instructions. As a result, the model may respond with short answers even when explicitly instructed to provide explanations.

To illustrate this phenomenon, we evaluate the reasoning capability of LVLMs by systematically modifying existing evaluation datasets. Specifically, we transform answer directives into justification requests by replacing phrases such as \textit{``Answer in one letter only''} or \textit{``Answer as succinctly as possible''} with \textit{``Justify your answer.''} We use \textit{Claude Sonnet~4.5} as an automated evaluator, following the \textit{scoring evaluation} protocol of \citet{chen2024mllm}. Each model response is evaluated along five dimensions: \textbf{Correctness}, a binary measure of whether the chosen answer matches the ground truth; \textbf{Presence}, whether the response includes a genuine justification beyond a bare answer; \textbf{Relevance} (0--5), how directly the explanation addresses the question and chosen answer; \textbf{Faithfulness} (0--5), how well the reasoning is grounded in the given image and prompt while penalizing hallucinations; and \textbf{Verbosity} (0--5), penalizing unnecessary length. The judge is provided with a system prompt and receives the original task instruction, image, ground-truth answer, and candidate model output for each evaluation sample, see Appendix for details. Due to limited token budget, up to 350 samples per dataset are randomly selected for evaluation. The overall quality score is computed as
\begin{align}
    \text{Overall} = \text{Faithfulness} + \text{Relevance} - \text{Verbosity} / 2. \nonumber
\end{align}

Results are summarized in Table~\ref{tab:explain_summary}, comparing our method with the VILA-U and VILA baselines, as well as continuous fine-tuning methods, including LDIFS~\citep{mukhoti2023fine}, Projector FT \citep{zhang2024visually}, and LLM LoRA \citep{hu2022lora}. All continuously fine-tuned models suffer from catastrophic forgetting of instruction-following capabilities, as reflected in the low reasoning presence and overall reasoning quality scores. In contrast, \textsc{CRAFT} maintains balanced performance across all metrics, improving both accuracy and reasoning capability over the baselines.

\subsection{Decoupling Vision and Language}
CRAFT genuinely improves the visual representations learned by the adapted encoder, allowing LVLMs with very different language backbones to perform better in their target domains. Table~\ref{tab:crossllm} further supports this through a detailed study of CRAFT’s modularity. We fine-tune the encoder using two surrogate models at opposite ends of the size spectrum, Qwen2-0.5B and VILA-U-7B, and run inference on five backbones that vary in size and architecture.
Across these backbones, CRAFT delivers consistent and often substantial improvements. For example, when the Qwen2.5-3B backbone uses an encoder trained with the Qwen2-0.5B surrogate, its average accuracy increases from 46.74\% in the zero-shot setting to 59.98\%. Another interesting observation is that training with a stronger surrogate does not always yield the best results. For example, on the Qwen2.5-0.5B backbone, the encoder trained with the Qwen2-0.5B surrogate outperforms the one trained with the VILA-U-7B surrogate. We suspect this is because Qwen2 and Qwen2.5 share closer architectural designs than VILA-U, so they interpret visual tokens in more similar ways, making the adapted encoder transfer more smoothly. Additional results are provided in the Appendix.

\begin{table}[t]
    \centering
    \small
    \setlength{\tabcolsep}{7pt}
    \renewcommand{\arraystretch}{1.16}
    \begin{adjustbox}{max width=\linewidth}
    \begin{tabular}{llcc}
        \toprule
        &  \multicolumn{2}{c}{\textbf{Training Efficiency}} & \\
        \midrule
        \textbf{Surrogate} & \textbf{Method} & \textbf{VRAM (MiB)} & \textbf{Train time (min)} \\
        \midrule
        VILA-7B & LoRA FT           & 27,725  & 5.09  \\
        VILA-7B & Vision FT         & 27,671  & 5.19  \\
        VILA-7B & Projector FT      & 27,383  & 4.46  \\
        Qwen2-0.5B & \textbf{Ours}           & 10,678  & 1.35 \\
        VILA-U-7B & \textbf{Ours}            & 20,754  & 2.27 \\
        \midrule
        & \multicolumn{2}{c}{\textbf{Inference Efficiency (Pruning)}} & \\
        \midrule
        \textbf{Backbone} & \textbf{Method} & \textbf{TFLOPs} & \textbf{Runtime (ms)}\\
        \midrule
        VILA-U-7B & Zero-shot & 2.56 & 52.40 $\pm$ 1.95 \\
        VILA-U-7B & \textbf{Ours} & 2.15 & 48.92 $\pm$ 2.02 \\
        \bottomrule
    \end{tabular}
    \end{adjustbox}
    \caption{
        Training and inference compute statistics. CRAFT training is lightweight when using a small surrogate, requiring substantially less memory and runtime compared with VILA-7B fine-tuning. With a keep ratio of 0.8, CRAFT’s pruning achieves meaningful reductions in both FLOPs and inference latency.
    }
    \label{tab:compute}
\end{table}

\begin{table}[t]
\small
\centering
\setlength{\tabcolsep}{5pt}
\renewcommand{\arraystretch}{1.15}
\begin{adjustbox}{max width=0.98\linewidth}
\begin{tabular}{l|cccc|c}
\toprule
Setting             & VQARAD & Dogs & PlantVillage & IconQA & All-Avg \\
\midrule
w.o. $\mathcal{L}_{\text{con}}$      & 45.13 & 71.57 & 45.69 & 47.24 & 52.41 \\
w.o. $\mathcal{L}_{\text{commit}}$   & 10.33 & 16.53 & 25.97 & 3.31  & 14.04 \\
w.o. $\mathcal{L}_{\text{SAL}}$      & 37.87 & 83.66 & 75.03 & 15.49 & 53.01 \\
\textbf{Ours (All)}   & \textbf{45.67} & \textbf{84.77} & \textbf{77.27} & \textbf{48.50} & \textbf{64.05} \\
\bottomrule
\end{tabular}
\end{adjustbox}
\caption{Loss ablation on the proposed training objectives. Each row removes one loss component. Dropping any component degrades performance to varying degrees, highlighting the contribution of each part of the objective. All encoders are trained and evaluated using the VILA-U-7B backbone.}
\label{tab:loss_ablation}
\end{table}

\subsection{Runtime and Efficiency}
Table~\ref{tab:compute} reports the training and inference efficiency of models under the same experimental settings. Training efficiency is measured on the \textit{Dog} dataset using eight 40GB A100 GPUs with an effective batch size of 32. Inference efficiency is averaged over 1000 runs with a batch size of 1 across ten datasets.
CRAFT training becomes especially lightweight when the surrogate model is small. For example, using Qwen2-0.5B as the surrogate cuts memory usage by 61.6\% and reduces training time by 73.5\% compared with other fine-tuning methods on VILA-7B, while still delivering competitive performance as shown in Table~\ref{tab:explain_summary}. At inference time, CRAFT benefits from the discrete structure of visual tokens by removing redundant tokens to reduce computation. With a keep ratio of about 0.8, token pruning lowers FLOPs by 16\% and reduces runtime by 7\% on average across datasets.

\begin{figure}[t]
\centering
\includegraphics[width=1.0\linewidth]{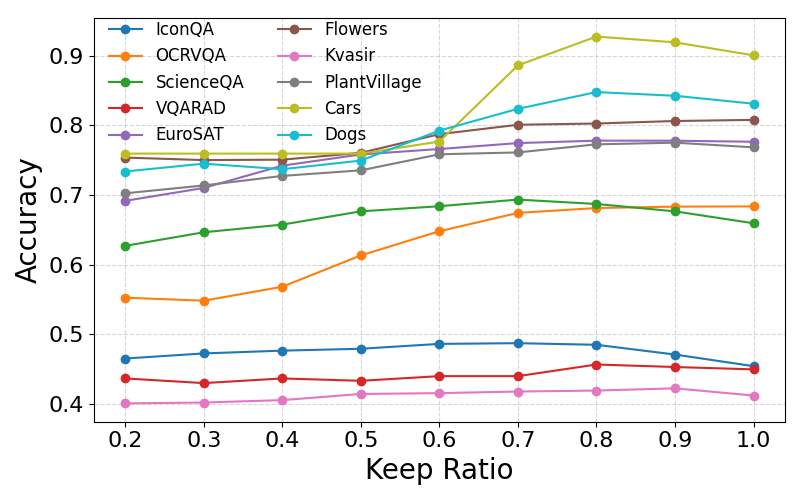}
\caption{Accuracy of CRAFT encoder with VILA-U-7B backbone on various datasets versus the keep ratio. Each curve represents a different dataset. The keep ratio is the ratio between target budget $M$ and image token number $N$; 1.0 indicates no pruning. Performance is consistently reliable when the keep ratio is above 0.6.}
\label{fig:ablation}
\end{figure}

\subsection{Ablation Studies}

\noindent\textbf{Ablation on Loss Components.} Table~\ref{tab:loss_ablation} shows the results of removing each of the three loss components. 
When the commitment loss is removed, performance drops sharply, likely because the visual features are no longer aligned with the codebook, causing the quantization step to damage the learned representations. 
The contrastive and SAL losses complement each other depending on the task. 
Removing the SAL loss lowers accuracy by 7.8\% and 33.0\% on \textit{VQARAD} and \textit{IconQA}, and removing the contrastive loss decreases accuracy by 13.2\% and 31.6\% on \textit{Dogs} and \textit{PlantVillage}. 
These results suggest that the contrastive loss helps more on tasks that rely on direct visual evidence, such as recognition and classification, and the SAL loss helps more on tasks that need deeper text understanding and reasoning.

\noindent\textbf{Ablation on Keep Ratio.}
By removing redundant and task-irrelevant tokens at inference, the model can focus on the most informative regions of the image, which improves performance. Figure~\ref{fig:ablation} shows how accuracy changes with different keep ratios across datasets. The effect of pruning varies a lot: datasets such as \textit{Kvasir} and \textit{VQARAD} show little accuracy drop even when only 20\% of tokens are kept, while \textit{Dogs} and \textit{Cars} degrade more noticeably. This difference reflects how information is distributed: medical VQA often depends on a few localized patches (as in the middle example of Figure~\ref{fig:teaser}), whereas fine-grained classification relies on broader spatial details, making more tokens important. Empirically, light pruning with a keep ratio above 0.6 is consistently reliable, and we use 0.8 throughout this paper.
We also ablate the components of the pruning strategy itself at a fixed keep ratio of 0.8. Random token selection achieves 62.10\% average accuracy. Adding rarity-weighted per-ID quotas improves this to 63.55\%, confirming that frequency-based budget allocation is beneficial. Further incorporating quantization-residual ranking within each ID raises accuracy to 63.86\%, and the full strategy with spatial isolation scoring reaches 64.05\%. Each component contributes incrementally, with the rarity-weighted allocation providing the largest single gain. Additionally, replacing the per-dataset frequency prior $p_{\text{dom}}(k)$ with one estimated from ImageNet-1K yields comparable performance (64.01\%), suggesting that the pruning strategy is robust to the choice of reference corpus.

\begin{table}[t]
\small
\centering
\setlength{\tabcolsep}{5pt}
\renewcommand{\arraystretch}{1.15}
\begin{adjustbox}{max width=0.98\linewidth}
\begin{tabular}{c|ccccc|c}
\toprule
Codebook Size 
& Dogs & Cars & Flowers & PlantVillage & IconQA & Avg \\
\midrule
10\%    
& 28.00 & 30.93 & 29.04 & 43.83 & 29.59 & 32.28 \\
25\%    
& 40.33 & 29.54 & 39.75 & 49.02 & 33.74 & 38.48 \\
50\%    
& 61.30 & 64.58 & 61.74 & 62.12 & 40.66 & 58.08 \\
100\%   
& 84.77 & 92.74 & 80.26 & 77.27 & 48.50 & 76.71 \\
\bottomrule
\end{tabular}
\end{adjustbox}
\caption{
Effect of codebook size on performance across benchmarks. Larger codebook size consistently improves accuracy, emphasizing the importance of both sufficient codebook capacity and a capable vision encoder for robust LVLM performance.
}
\label{tab:codebook_ablation}
\end{table}

\noindent\textbf{Ablation on Codebook Size.}
A primary limitation of discrete codebooks is their limited expressiveness in representing image features. In VILA-U, the default vision-token codebook contains 16384 entries. To assess the effect of codebook capacity, we uniformly downsample the codebook to 50\%, 25\%, and 10\% of its original size and apply CRAFT using the VILA-U-7B backbone for both training and evaluation. As shown in Table~\ref{tab:codebook_ablation}, larger codebooks provide stronger representational capacity, leading to higher overall accuracy. Specifically, the average accuracy increases from 32.28\% at 10\% capacity to 76.71\% at full size, clearly demonstrating the significance of the encoder’s capacity in LVLM performance.

\section{Future Work}
For future work, we would like to understand how changes to the shared codebook affect backward compatibility. While the codebook is assumed to be fixed in this paper, in practice, future models may introduce expanded or more detailed visual vocabularies~\citep{qu2025tokenflow,ma2025unitok,long2025domain}. If new entries are added while retaining the old ones, an open question is whether existing vision encoders will continue to function properly without additional retraining.

\section{Conclusion}
In this work, we introduced CRAFT, a lightweight and modular framework for LVLM specialization without altering the language model. CRAFT trains only the vision encoder while keeping a fixed discrete codebook, so the encoder learns to produce visual tokens that any LVLM sharing the same codebook can understand. This avoids modifying the language model and keeps its reasoning abilities intact.
Through comprehensive experiments across a diverse set of visual and reasoning benchmarks, we demonstrated that CRAFT significantly improves domain-specific classification and VQA accuracy by good margins, while maintaining faithful reasoning capability. The shared codebook interface allows encoders fine-tuned with small surrogate models to transfer seamlessly to larger LLMs with different architectures, delivering strong results at a fraction of the training and inference cost and offering a practical solution for resource-constrained settings.

%% file: sec/X_suppl.tex
\clearpage
\setcounter{page}{1}
\maketitlesupplementary

\section{More Details on LLM Judge Experiments}
\label{appendix:more_judge_details}

We use \textit{Claude Sonnet 4.5} as an automated judge to evaluate the quality of vision-language model responses, following the \textit{scoring evaluation} approach from \citet{chen2024mllm}. The judge model analyzes both the correctness of the answer and the quality of the explanation provided by the evaluated model. The evaluation framework utilizes \textit{Claude-Sonnet-4-5} as the judge model. The judge model is configured with a maximum token limit of 1024 and a temperature of 0.6 to balance consistency with necessary variability.

The judge evaluates each model's response along five distinct dimensions: \textbf{Correctness}, a binary metric that determines whether the assistant's chosen answer is correct with respect to the ground-truth answer. The judge allows for synonymous answers, recognizing that multiple phrasings may represent the same correct information. The second dimension is \textbf{Presence}, also binary, which assesses whether the response includes a genuine justification rather than only a bare answer. A response passes this criterion if it contains explanations that reference evidence beyond the final answer. The third dimension is \textbf{Relevance}, measured on a 0--5 scale that quantifies how well the explanation addresses the user's question and relates to the chosen answer. A score of 0 indicates the response is off-topic, while a score of 5 indicates it directly addresses the question. The fourth dimension is \textbf{Faithfulness}, also on a 0--5 scale, which evaluates how well the explanation is grounded in the given image and prompt. This dimension penalizes hallucinations and contradictions, with 0 representing hallucinated content and 5 representing full consistency with observable evidence. The fifth dimension is \textbf{Verbosity}, scored on a 0--5 scale that penalizes unnecessary verbosity relative to the substance of the response. Note that lower scores indicate better verbosity: a score of 0 means the response is concise, while a score of 5 indicates it is overly long.

The judge receives a carefully crafted system prompt that explicitly defines each evaluation dimension and requires the judge to output its assessment in a strict JSON format containing the five dimensions, as well as a short two sentence justification providing the key rationale for the scores. The JSON format encodes ``correctness'' and ``presence'' as boolean values, ``relevance'', ``faithfulness'', and ``verbosity'' as integers in their respective ranges, and the final justification as a string. For each evaluation sample, the judge receives a structured user prompt containing the original task instruction from the evaluation dataset, the image, all available options if the task is multiple-choice, the ground-truth answer provided for orientation, and the candidate model's answer and explanation. The system prompt is designed as follows:
\begin{tcolorbox}[colback=gray!5!white, colframe=black!60!black, title=System Prompt for Claude 4.5]
\textbf{You are a careful, unbiased multimodal judge proficient in analyzing vision reasoning problems.}

You will be provided with the task prompt, possible options (if present), the ground-truth answer, and the task image.

Judge each assistant’s response along five dimensions:

\textbf{Correctness (binary):} Is the assistant’s chosen answer correct with respect to the ground-truth? (Allow for synonymous or equivalent answers.)

\textbf{Presence (binary):} Does the response include a genuine justification rather than only a bare answer? (Look for explanations that reference evidence, reasoning, or why.)

\textbf{Relevance (0--5):} How well does the explanation address the user’s instruction/question and the chosen answer? 0 means off-topic, 5 means directly and fully addresses both.

\textbf{Faithfulness (0--5):} How well is the explanation grounded in the given image and prompt (and options)? 0 means hallucinated or contradictory; 5 means fully consistent with observable evidence and the question context.

\textbf{Verbosity (0--5):} Penalize unnecessary verbosity; 0 means succinct and complete, 5 means overly long or repetitive relative to substance.

\vspace{0.5em}
\noindent\textbf{Output strictly in the following JSON:}

\textit{
  "correctness": true or false,
  "presence": true or false,
  "relevance": 0-5,
  "faithfulness": 0-5,
  "verbosity": 0-5,
  "notes": "At most 2 short sentences explaining your scores."
}

\end{tcolorbox}

\begin{figure*}[ht]
    \centering
    \includegraphics[width=\linewidth]{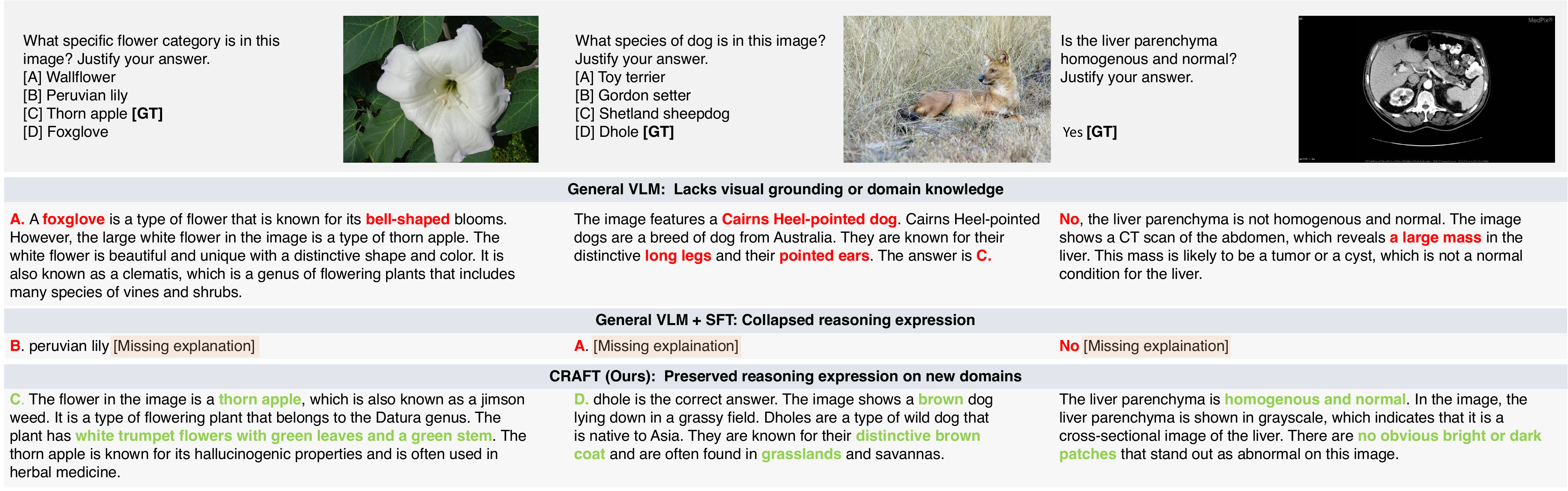}
    \caption{
        Additional examples following Figure~2 of the main paper. The general VLM makes incorrect predictions because errors from the vision encoder mislead the LLM: it misinterprets the flower’s shape, it hallucinates on the dog image when the encoder fails to extract reliable visual information, and it makes an incorrect medical judgement when the encoder does not have the necessary contextual understanding and instead provides misleading mass-like features. The fine-tuned baseline suffers from catastrophic forgetting of its instruction-following ability and cannot provide explanations even when explicitly asked. Our method delivers accurate predictions while also preserving more precise reasoning capability across these tasks.
    }
    \label{fig:teaser_v2}
    \vspace{-2mm}
\end{figure*}

\begin{figure*}[ht]
    \centering
    \includegraphics[width=0.64\linewidth]{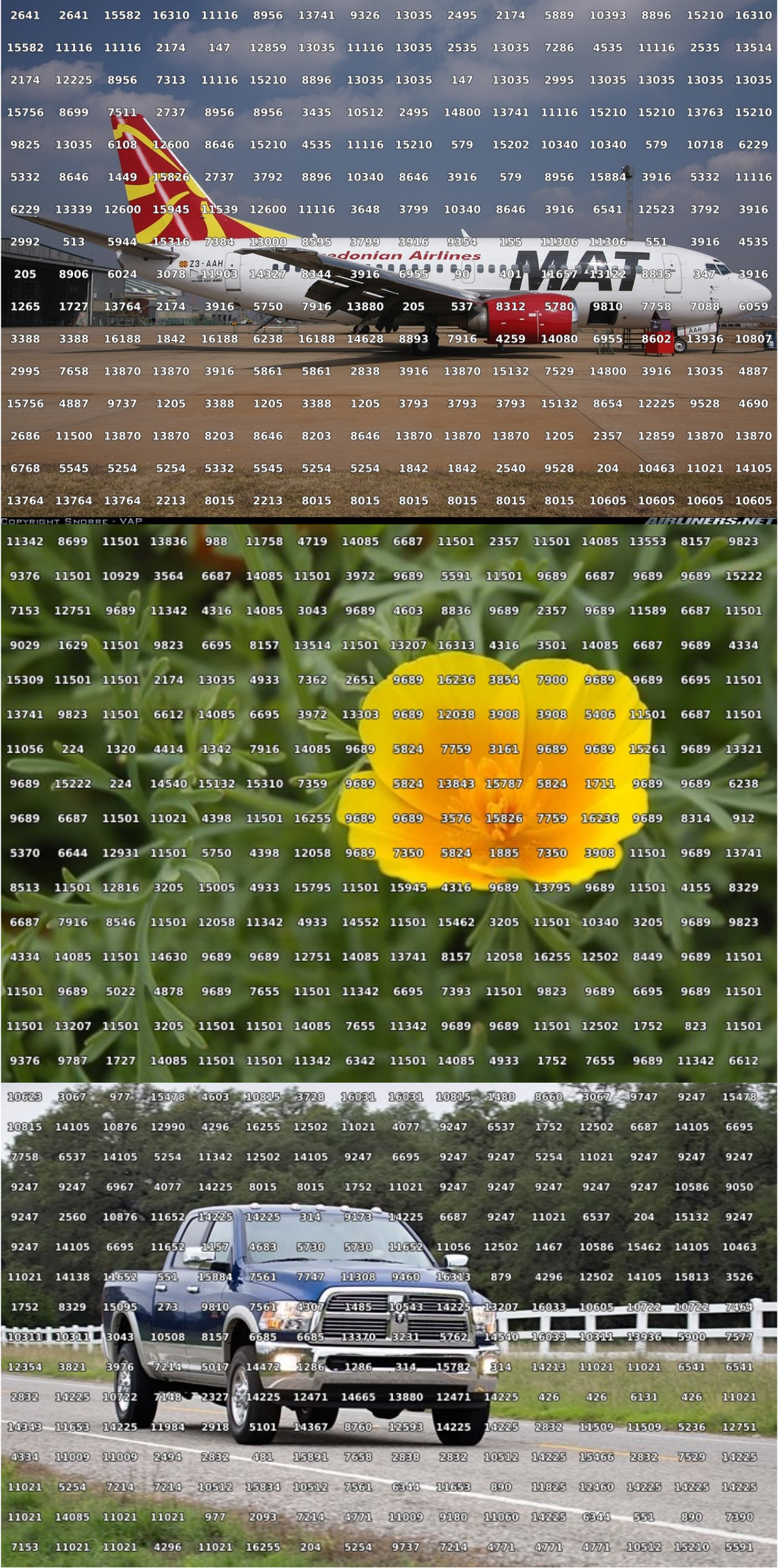}
    \caption{Images with the first-level residual quantization indices overlaid, following the VILA-U paper. We observe that homogeneous regions with redundant visual information often produce many repeated tokens sharing the same first-level indices, for example, 13035, 3793, and 8015 in the aircraft image; 11501 and 14085 in the flower image; and 9247 and 14225 in the car image. This repetition suggests that these areas can be pruned to give a more concise visual representation.}
    \label{fig:teaser_appendix}
\end{figure*}

\section{Surrogate and LVLM Training}

CRAFT works because all models share the same visual codebook. This codebook acts like a common “visual language” that both the vision encoder and the LLMs understand. When different LLMs use this shared codebook, a vision encoder trained with one model can work with another model immediately, without any extra alignment. Since the public VILA-U~\citep{wu2024vila} release includes only one 7B model based on Llama~2, to support our idea and create the experimental setup for this paper, we follow their training recipe to build more LVLMs using Qwen2 and Qwen2.5 backbones with sizes ranging from 0.5B to 3B. All of these models use the same VILA-U vision encoder and the same frozen codebook.

Following \citet{wu2024vila}, we perform a two stage training process. First, we train only the projector on samples from the LLaVA Visual Instruct CC3M Pretrain 595K dataset~\cite{liu2023visual} to provide a preliminary understanding of images. Afterwards, we train the projector and LLM with 1M image–text pairs from ShareGPT4V~\cite{chen2024sharegpt4v} and 150K instruction-following examples from LLaVA~\cite{li2024llavaonevision}. These datasets teach the surrogate models to understand and process the discrete visual tokens produced by the shared codebook. We select several open-source LLMs, including Qwen2-0.5B, Qwen2-1.5B, Qwen2.5-0.5B, Qwen2.5-3B~\cite{bai2025qwen2}, to serve as surrogate models during encoder adaptation. The Qwen models are pure language models, so we attach the VILA-U vision encoder to them and train them to work with the discrete visual tokens. During training, the VILA-U vision encoder and its codebook stay frozen. Only the projector and the LLM weights are updated. We train each surrogate for one epoch using AdamW with a learning rate of $10^{-4}$, a warmup ratio of 0.03, cosine decay, and no weight decay. The maximum sequence length is 8192 tokens. Training uses BF16 precision, DeepSpeed ZeRO-2, and gradient checkpointing on 8 A100 GPUs (40GB). We repeat this process for every LLM backbone, building a family of LVLMs that all speak the same discrete visual language.

After training, each surrogate model can guide the encoder during CRAFT. More importantly, this training is a one-time effort: an encoder adapted with one surrogate (such as Qwen2-0.5B) can be used directly with any other LLM that shares the same codebook, including VILA-U-7B or Qwen2.5-3B. No extra training or alignment is needed. We hope our findings encourage more work on unified visual codebooks so that future research can build on a larger set of shared resources and models.

\begin{table*}[t]
\centering
\small
\setlength{\tabcolsep}{4.5pt}
\renewcommand{\arraystretch}{1.15}
\begin{adjustbox}{max width=\linewidth}
\begin{tabular}{l|l|l|cccccccccc}
\hline
Method & Train Surrogate & Inference Backbone & IconQA & OCRVQA & ScienceQA & VQARAD & EuroSAT & Flowers & Kvasir & PlantVillage & Cars & Dogs \\
\hline
Zero-shot & - & Qwen2-0.5B & 2.40 & 55.43 & 43.43 & 36.67 & 61.70 & 42.84 & 28.74 & 28.23 & 35.47 & 35.17 \\
\textbf{Ours} & Qwen2-0.5B & Qwen2-0.5B & \textbf{30.97} & 54.33 & 47.24 & 39.00 & \textbf{80.68} & 52.29 & \textbf{38.11} & 76.43 & \textbf{57.94} & \textbf{67.17} \\
\textbf{Ours} & Qwen2.5-0.5B & Qwen2-0.5B & 21.37 & 53.73 & 47.84 & 39.33 & 80.41 & 52.53 & 34.75 & 62.70 & 56.47 & 62.57 \\
\textbf{Ours} & Qwen2-1.5B & Qwen2-0.5B & 21.27 & 53.00 & 47.64 & \textbf{41.33} & 78.65 & 50.97 & 32.08 & \textbf{80.10} & 54.57 & 60.87 \\
\textbf{Ours} & Qwen2.5-3B & Qwen2-0.5B & 20.63 & 53.97 & \textbf{47.94} & 40.00 & 75.37 & \textbf{53.83} & 35.68 & 75.57 & 55.20 & 65.31 \\
\textbf{Ours} & VILA-U-7B & Qwen2-0.5B & 17.73 & \textbf{60.27} & 47.44 & 37.33 & 70.70 & 51.44 & 31.85 & 69.67 & 57.70 & 54.77 \\
\hline
Zero-shot & - & Qwen2.5-0.5B & 1.57 & 41.93 & 55.63 & 36.67 & 67.46 & 50.89 & 32.68 & 27.63 & 31.37 & 34.60 \\
\textbf{Ours} & Qwen2-0.5B & Qwen2.5-0.5B & 26.13 & 43.70 & 56.91 & 42.67 & 81.46 & 61.88 & 46.68 & 70.27 & 50.10 & 65.54 \\
\textbf{Ours} & Qwen2.5-0.5B & Qwen2.5-0.5B & \textbf{31.07} & 43.00 & 56.71 & \textbf{46.33} & \textbf{83.48} & \textbf{62.55} & \textbf{51.90} & 67.43 & \textbf{50.94} & \textbf{71.44} \\
\textbf{Ours} & Qwen2-1.5B & Qwen2.5-0.5B & 27.13 & 43.60 & 56.86 & 43.33 & 81.37 & 56.78 & 47.96 & \textbf{80.87} & 46.60 & 66.01 \\
\textbf{Ours} & Qwen2.5-3B & Qwen2.5-0.5B & 25.43 & 43.57 & \textbf{57.06} & 41.33 & 79.76 & 61.77 & 49.93 & 72.10 & 41.64 & 67.77 \\
\textbf{Ours} & VILA-U-7B & Qwen2.5-0.5B & 19.27 & \textbf{51.73} & 56.91 & 41.00 & 70.26 & 57.10 & 43.67 & 65.97 & 43.97 & 57.97 \\
\hline
Zero-shot & - & Qwen2-1.5B & 7.07 & 59.33 & 68.27 & 37.00 & 68.20 & 55.15 & 33.26 & 51.17 & 55.27 & 55.90 \\
\textbf{Ours} & Qwen2-0.5B & Qwen2-1.5B & 22.93 & 59.47 & 74.11 & 42.67 & 82.87 & 69.48 & 44.37 & 82.53 & 73.54 & 80.54 \\
\textbf{Ours} & Qwen2.5-0.5B & Qwen2-1.5B & 26.73 & 58.77 & 73.96 & 41.67 & 85.04 & 69.97 & 37.53 & 69.13 & 73.87 & 80.57 \\
\textbf{Ours} & Qwen2-1.5B & Qwen2-1.5B & \textbf{27.50} & 60.43 & 74.06 & 42.67 & \textbf{87.30} & 71.25 & \textbf{44.71} & \textbf{87.40} & 77.67 & 81.31 \\
\textbf{Ours} & Qwen2.5-3B & Qwen2-1.5B & 19.53 & 60.27 & 73.96 & \textbf{44.33} & 80.37 & \textbf{74.80} & 41.47 & 77.90 & 75.80 & \textbf{81.91} \\
\textbf{Ours} & VILA-U-7B & Qwen2-1.5B & 22.90 & \textbf{68.67} & \textbf{74.41} & 41.33 & 80.68 & 68.44 & 34.29 & 81.37 & \textbf{85.87} & 78.24 \\
\hline
Zero-shot & - & Qwen2.5-3B & 3.80 & 58.13 & 63.26 & 33.67 & 66.46 & 59.16 & 22.25 & 46.73 & 63.07 & 50.90 \\
\textbf{Ours} & Qwen2-0.5B & Qwen2.5-3B & 19.80 & 59.63 & 63.60 & 36.33 & 84.11 & 65.76 & 47.15 & 81.97 & 70.60 & 70.84 \\
\textbf{Ours} & Qwen2.5-0.5B & Qwen2.5-3B & 22.83 & 59.60 & \textbf{63.80} & 39.00 & 79.11 & 66.03 & 36.14 & 75.40 & 69.60 & 68.14 \\
\textbf{Ours} & Qwen2-1.5B & Qwen2.5-3B & 25.67 & 59.57 & 63.55 & \textbf{41.00} & \textbf{89.67} & 67.20 & 48.42 & \textbf{89.80} & 71.74 & 71.74 \\
\textbf{Ours} & Qwen2.5-3B & Qwen2.5-3B & \textbf{32.60} & 60.07 & 63.50 & 40.33 & 85.89 & \textbf{73.28} & \textbf{50.62} & 71.13 & \textbf{81.14} & \textbf{75.77} \\
\textbf{Ours} & VILA-U-7B & Qwen2.5-3B & 25.03 & \textbf{69.17} & 63.70 & 39.67 & 80.50 & 66.02 & 38.69 & 81.37 & 77.64 & 67.41 \\
\hline
Zero-shot & - & VILA-U-7B & 10.17 & 66.40 & 65.59 & 35.67 & 69.15 & 75.80 & 29.20 & 43.83 & 72.50 & 82.40 \\
\textbf{Ours} & Qwen2-0.5B & VILA-U-7B & 30.50 & 66.77 & 68.86 & 39.00 & 78.87 & 72.31 & 32.55 & 70.70 & 78.80 & 74.11 \\
\textbf{Ours} & Qwen2.5-0.5B & VILA-U-7B & 31.90 & 65.90 & 68.81 & 42.33 & 75.76 & 72.18 & 29.65 & 57.53 & 76.70 & 72.01 \\
\textbf{Ours} & Qwen2-1.5B & VILA-U-7B & 33.30 & 66.80 & \textbf{69.15} & 42.33 & \textbf{83.11} & 74.10 & 29.19 & \textbf{83.43} & 77.54 & 77.94 \\
\textbf{Ours} & Qwen2.5-3B & VILA-U-7B & 33.73 & 67.97 & 69.06 & \textbf{46.33} & 79.70 & 77.92 & 33.71 & 66.70 & 87.34 & 83.81 \\
\textbf{Ours} & VILA-U-7B & VILA-U-7B & \textbf{48.50} & \textbf{68.13} & 68.71 & 45.67 & 77.80 & \textbf{80.26} & \textbf{41.93} & 77.27 & \textbf{92.74} & \textbf{84.77} \\
\hline
\end{tabular}
\end{adjustbox}
\caption{Cross-LLM transfer with a shared discrete codebook. A vision encoder trained with one surrogate language model during training can be directly paired with different LLM backbones at test time. The adapted encoder consistently improves the performance of five LLM backbones across ten datasets, demonstrating that CRAFT supports portable vision encoder fine-tuning without requiring re-alignment of language models.}
\label{tab:crossllm_appendix}
\vspace{-2mm}
\end{table*}

\section{Quantization Details}

Following VILA-U~\cite{wu2024vila}, each image is discretized using a multi-level residual vector quantizer. For an input image $x$, the vision encoder $E_\theta$ outputs a grid of continuous patch features
\[
z = E_\theta(x) \in \mathbb{R}^{N \times d},
\]
where each $z_i$ is a $d$-dimensional feature for patch $i$.

VILA-U applies residual quantization in four stages. At the first level, each $z_i$ is mapped to its nearest codeword in the first-level codebook:
\[
c^{(1)}_{k^{(1)}_i} = \arg\min_{c \in \mathcal{C}^{(1)}} \| z_i - c \|_2,
\qquad
r^{(1)}_i = z_i - c^{(1)}_{k^{(1)}_i}.
\]
The residual is then quantized again:
\[
c^{(2)}_{k^{(2)}_i} = \arg\min_{c \in \mathcal{C}^{(2)}} \| r^{(1)}_i - c \|_2,
\qquad
r^{(2)}_i = r^{(1)}_i - c^{(2)}_{k^{(2)}_i}.
\]
This procedure repeats for levels $\ell = 3,4$, producing four indices
\[
(k^{(1)}_i,\; k^{(2)}_i,\; k^{(3)}_i,\; k^{(4)}_i)
\]
for each patch. The final quantized feature is obtained by summing all level-wise codewords:
\[
\tilde{z}_i = \sum_{\ell=1}^{4} c^{(\ell)}_{k^{(\ell)}_i}.
\]
These four discrete tokens per patch are then projected into the LLM embedding space.

For visualization, Figure~4 in the main paper shows only the subsampled first-level indices $k^{(1)}_i$, with additional examples in Figure~\ref{fig:teaser_appendix}. The pruning algorithm also operates on these first-level assignments. During scoring, we compute
\[
e_i = \| z_i - c^{(1)}_{k^{(1)}_i} \|_2^2,
\]
so each patch is ranked by how well its continuous feature matches its first-level codeword. Intuitively, the index $i$ correlates to the scale at which we perform discretization. As $i$ increases, the value of the residual progressively decreases. Thus, the different scales of discretization allow one to capture different types of information. This allows for greater expressivity in the codebook, as certain codebook entries can represent coarse grained concepts (\textit{e.g.}, car, dog) while other entries capture fine-grained details (\textit{e.g.}, texture, lighting). 

Throughout the main text, expressions such as $c_k$ or ``the codebook index $k$’’ refer by default to the first-level codeword $c^{(1)}_{k^{(1)}}$. This keeps the notation concise: although VILA-U assigns four indices to each patch, the first-level codes are the most interpretable and are the ones visualized in Figures~4 in the main paper and Figure~\ref{fig:teaser_appendix}. The full four-level quantized representation is still used during training and inference.

\section{More Architectural Details}
In this work, we build upon the VILA-U~\cite{wu2024vila} 7B DLVLM. The discrete image encoder is initialized with SigLIP~\cite{zhai2023sigmoid} weights and is additionally trained with language alignment, reconstruction loss, and commitment loss on a large, general purpose image dataset. The embeddings of the penultimate layer are discretized via nearest-neighbor search in the codebook with a progressive residual quantization process. 
After the discrete visual encoder is trained, it is aligned with an off-the-shelf 7B LLM (Llama 2) by tuning on 8M image-text pairs. We refer readers to the original work for greater detail. We selected the VILA-U model due to its open-source nature, and emphasize that CRAFT has no reliance on any VILA-U specific components, depending only upon the codebook which is present in all DLVLMs.

\section{Datasets}
We provide greater information on the following datasets used to evaluate CRAFT:

\begin{itemize}
    \item \textbf{IconQA}~\citep{lu2021iconqa}: Reasoning over computer generated icons to test abstract diagram recognition, visual reasoning skills, and common sense. The multiple choice split consists of 18,946 samples with 6,316 train and validation samples. 
    \item \textbf{OCRVQA}~\citep{mishraICDAR19}: Tests the model's ability to recognize text in images and answer questions with that information. The dataset contains 166k training samples and 20k validation and testing samples. In our experiments, we subsampled 30k out of the 166k training data for training.
    \item \textbf{ScienceQA}~\citep{lu2022learn}: Evaluates the ability to answer scientific questions in context of a provided image, often requiring prior background knowledge of the scientific principles. Dataset contains roughly 21k samples. 
    \item \textbf{VQARAD}~\citep{lau2018dataset}: Short answer dataset with medical radiology scans with about 2k training samples and 450 testing samples. VQARAD represents a highly specialized domain with limited training data. 
    \item \textbf{Kvasir}~\citep{Pogorelov:2017:KMI:3083187.3083212}: Classification dataset consisting of gastrointestinal disease images. The images are grouped into 8 classes, each with 500 samples. 
    \item \textbf{EuroSAT}~\citep{helber2019eurosat}: Satellite imagery classification dataset with 10 classes and a total of 27000 samples. EuroSAT represents a new domain that the LVLMs are unlikely to encounter during the pretraining process. 
    \item \textbf{Flowers}~\citep{nilsback2006visual}: 102 different flower classes, with 40 to 258 samples per class. While the LVLM has doubtless encountered flowers during pretraining, Flowers-102 is a fine-grained visual recognition dataset that requires distinction between different flower species. 
    \item \textbf{PlantVillage}~\citep{salathe2016plantvillage}: Contains 54303 leaf images grouped into 38 classes by species and disease. PlantVillage tests both domain-shift (minimal representation during pretraining), as well as fine-grained understanding (detailed visual understanding required to differentiate leaf diseases and species). 
    \item \textbf{Cars}~\citep{krause20133d}: The Stanford-Cars dataset contains 16,185 total images of 196 classes of cars. It evaluates fine-grained visual recognition capabilities. 
    \item \textbf{Dogs}~\citep{khosla2011novel}: The Stanford-Dogs dataset contains 120 dog breeds across 20,580 total images and also evaluates fine-grained visual capabilities. 
\end{itemize}

\noindent\textbf{Dataset Preprocessing.}
For the classification-style datasets, they are transformed into VQA datasets for compatibility with the LVLM. Following FOCI~\cite{geigle2024african}, we choose to format them as four-option multiple choice tasks to prevent ambiguity in the answer resulting from incorrect formatting or different levels of specificity. In addition, given that LVLMs are world models and can reason over labels not present during training, we omit 20\% of the label classes during training, allowing us to test the ability of the model to generalize to novel labels. This is especially critical as LVLMs are notorious for overfitting to the training distribution.

\section{In-Depth Results Discussion}

\subsection{Main Paper Table 1 Baselines and Results}
In this table, we compare CRAFT fine-tuning to other forms of visual LVLM adaptation, including Vision FT, Projector tuning, and LDIFS. Vision FT represents the naive and computationally expensive method of fine-tuning the continuous vision encoder. The language model is not updated alongside the vision encoder, and can result in suboptimal understanding of the new visual embeddings. This phenomenon appears when Visual FT is performed on the EuroSAT and Dogs datasets, in which the accuracy after fine-tuning is \textit{lower} than before, demonstrating that visual fine-tuning of continuous LVLMs can corrupt the model. 

Projector fine-tuning is commonly employed in works that target object classification accuracy~\cite{zhang2024visually, geigle2024african}. These works argue that on some datasets, both the vision encoder and language model understand the objects within the image, yet insufficient data during the alignment stage results in a poor projection from image to language space. However, they typically include \textit{all the alignment data} during this projector fine-tuning stage to prevent overfitting, which is computationally expensive and impossible for closed-data models. For fairness and practicality, we evaluate only with the domain-specific dataset, discovering that projector fine-tuning leads to minimal improvement over the original model (1.26\%) on average. Similarly to Vision FT, we encounter severe degradation on certain datasets -- almost 10\% on Dogs and over 13\% on Flowers.

LDIFS proposed a method for mitigating embedding space shift when fine-tuning CLIP encoders. They focused on the domain of CLIP vision encoders (without LLMs), and found that imposing an $\ell_2$ regularization term could reduce feature drift. Since we hypothesize that the language model may not comprehend drastically altered visual embeddings, we evaluate our methods against LDIFS visual finetuning. Surprisingly, this method does worse when compared against both Vision FT and Projector FT, and underperforms the original model on average. It appears that this method may not be transferrable to LVLM vision encoder fine-tuning. 

Despite utilizing a surrogate model approach that is more lightweight (involves a smaller language model), CRAFT results in greater gains in accuracy when compared to the baselines. The best performing baseline, Vision-FT, elevates the accuracy by an average of 6.69\% over the original model. In contrast, the worst performing CRAFT variant achieves an average accuracy gain of 4.21\%, while the best CRAFT variant boasts an impressive 13.51\% improvement over the original discrete model. Additionally, with the exception of very small surrogate models on the Dogs dataset, CRAFT will not impair the model's accuracy.

\subsection{Main Paper Table 2 Baselines and Results}
Aside from testing performance on classification and VQA tasks, we also evaluate the reasoning capabilities of our models. We introduce an additional baseline of VILA-LLM-LoRA that modifies the weights of the language model. Despite attaining competitive correctness scores, we observe that VILA-LLM-LoRA exhibits a strong tendency to overfit to the syntax of its training data and fails to provide explanations. VILA-LLM-LoRA has by far the lowest presence score of all models, demonstrating that the LLMs language capabilities have collapsed. This result highlights the importance of \textit{visual} fine-tuning, which is less likely to collapse the language model. Nonetheless, fine-tuning the vision encoder can still result in diminishing language capabilities. We observe that VILA-LDIFS and VILA-Projector-FT have lower presence scores than the original model, showcasing that the fine-tuned projected visual tokens may have altered the behavior of the language model. CRAFT, with its discrete surrogate tuning approach, achieves competitive or superior presence, relevance, and faithfulness scores in comparison to the original discrete model, demonstrating that a discrete visual codebook offers unique advantages in retaining LLM functionality.

\subsection{Main Paper Table 3 Baselines and Results}
In Table 3, we showcase various combinations of LLM backbones and surrogates in which the surrogate is either smaller or larger than the backbone. We observe that surrogate fine-tuning will almost always outperform the original model. The only exception appears in high detailed datasets (fine-grained visual classification) when the gap between the surrogate and original language model is large (e.g., Qwen0.5B and VILA-U-7B). Additionally, the similarity between the surrogate and the inference backbone also has an impact on the resulting accuracy. Surrogates that are more similar to the inference backbones consistently outperform ones that are highly distinct. We provide a complete table in Table~\ref{tab:crossllm_appendix} of the appendix.

\section{Visual Codebook Indices}
In Figure~\ref{fig:teaser_appendix}, we showcase the selected codebook indices for selected images of airplanes, flowers and trucks. We can see that semantically irrelevant background information often shares the same token, providing motivation for our inference time token pruning strategy. For example, the 11021 token appears to relate to vegetation, appearing frequently on the grassy patches on the bottom left and middle right side of the truck image.